\documentclass[runningheads]{llncs}
\usepackage[T1]{fontenc}
\usepackage{graphicx,verbatim}
\usepackage{color}
\usepackage{hyperref}

\usepackage{amsmath}
\usepackage{amssymb}
\usepackage{booktabs}
\usepackage{multirow}

\makeatletter
\usepackage{cleveref}
\usepackage{marvosym}

\Crefname{figure}{Fig.}{Figs.}
\begin{document}
\title{Unified Multimodal Model for Brain MRI Imputation and Understanding}
% \titlerunning{Unified Multimodal Model}
% MICCAI submission
\author{Zhiyun Song\inst{1}\textsuperscript{\Letter} \and
Che Liu\inst{1}\textsuperscript{\Letter} \and
Tian Xia\inst{1} \and
Avinash Kori\inst{1} \and
Wenjia Bai\inst{1,2}}
% index{Song, Zhiyun; Liu, Che; Xia, Tian; Kori, Avinash; Bai, Wenjia}
%
\authorrunning{Z. Song et al.}
\institute{Department of Computing, Imperial College London, London, UK\and
Department of Brain Sciences, Imperial College London, London, UK\\
\email{z.song25, che.liu21@imperial.ac.uk}}

% \author{Anonymized Authors}  %% Added for anonymized MICCAI submission
% \authorrunning{Anonymized Author et al.}
% \institute{Anonymized Affiliations \\
%     \email{email@anonymized.com}}
%     \email{email@anonymized.com}}
  
\maketitle              % typeset the header of the contribution
\begin{abstract}
Multimodal large language models (MLLMs) hold great potential for medicine, as they inherit knowledge from LLM and allow multiple data modalities to be integrated, analysed and interpreted in natural language. However, the field of medical MLLMs is constrained by non-trivial challenges, notably the scarcity of high-quality training data and the frequent occurrence of missing data in the real-world clinical setting. Here, we propose a novel unified multimodal model, UniBrain, for brain magnetic resonance image (MRI) analysis. To address potential missing brain MRI modalities, we employ a unified training strategy to perform joint imaging modality imputation and brain image understanding. 
%Multi-parametric Magnetic Resonance Imaging (mpMRI) is essential for comprehensive brain disease diagnosis, yet real-world clinical workflows frequently suffer from missing modalities. Existing automated frameworks typically address this through disjointed imputation pipelines or implicit robust representations, failing to synergise visual generation with downstream diagnostic reasoning and often sacrificing clinical interpretability. In this paper, we propose Diagnosis with Generated Images (DGI), a novel framework built upon a Unified Foundation Model (UFM) to integrate missing modality imputation and clinical diagnosis within a single autoregressive process.
During training, an interleaved and description-enriched data flow is constructed to train the model in an autoregressive manner, enabling medical reasoning with generated multimodal data.
A self-alignment strategy is introduced to leverage dense image embeddings to learn fine-grained anatomical features without requiring detailed image captions. Furthermore, we propose a dynamic hidden state mechanism to alleviate the exposure bias during long-context multimodal inference. Extensive experiments on multi-disease brain MRI dataset demonstrate that UniBrain achieves high performance for brain image imputation, understanding, and disease diagnosis under various extents of modality incompleteness. Source code is publicly available at \url{https://github.com/zhiyuns/UniBrain}.
% On a dataset of 3,408 brain MRI scans with multiple modalities and disease types, extensive experiments demonstrate that UniBrain achieves high performance for brain image imputation, understanding, and disease diagnosis under varying extends of modality incompleteness.

% Authors must provide keywords and are not allowed to remove this Keyword section.

\end{abstract}
\section{Introduction}
Brain magnetic resonance imaging (MRI) plays an indispensable role in identifying structural abnormalities of the brain, diagnosing, and monitoring the progression of various neurological diseases.
%is a cornerstone of neuroimaging, with standard clinical protocols routinely acquiring multi-parametric sequences to provide a comprehensive view of brain pathology  \cite{multi_modality_significance1}. The complementary tissue contrasts across these modalities are essential for accurate diagnosis and treatment planning \cite{seiler2021multiparametric}.
Recently, multimodal large language models (MLLMs) have demonstrated potential in leveraging knowledge learnt from LLMs combined with modality-specific encoders to integrate and interpret multimodal data, including medical images, within a single model~\cite{alsaad2024multimodal,bercea2025nova,chen2024towards,yin2024survey}.
%fundamentally reshaped the landscape of medical image analysis \cite{chen2024towards,alsaad2024multimodal}, demonstrating remarkable capabilities in interpreting complex clinical scenarios by incorporating expert radiological knowledge and arbitrary modality inputs \cite{wu2026deep}.
However, developing MLLMs to interpret brain MRI scans faces significant challenges. Reading brain MRI scans requires radiological and neurological expertise, which may not be well captured by a general-purpose MLLM \cite{bercea2025nova}. Moreover, brain MRI can consist of multiple modalities, \emph{e.g.} T1-weighted, T2-weighted, FLAIR etc. Disease diagnosis would ideally require a comprehensive view of all modalities. In clinical reality, however, some modalities may be missing \cite{shen2019brain}, which creates the challenge for the MLLM to deal with the data modality gap.
% a pervasive challenge in real-world clinical practice is the frequent occurrence of missing MRI sequences due to scan time constraints, patient motion, or data corruption \cite{juvekar2024remind}. When deployed in these real-world settings, conventional MLLMs encounter a severe modality gap between their complete-data training distributions and incomplete inference data, leading to biased representations and degraded diagnostic accuracy.

To deal with missing modalities, existing literature generally follows two paradigms: \textit{explicit modality imputation}, and \textit{implicit representation learning}~\cite{wu2026deep}. 
Explicit methods leverage the available modalities to 
synthesise images of missing modalities, providing input for downstream tasks \cite{ResViT,rassmann2026regression,UniCOAL}. While these methods prioritise the fidelity of the synthesised images, imputation and subsequent analysis operate as a disjointed, two-stage pipeline, which might compromise the performance in learning downstream task-relevant representations from synthesised images. In contrast, implicit methods deal with missing modalities implicitly in the model. Some of these methods align the representations of the available and missing modalities within a shared space \cite{rahimpour2021cross,zhao2024deep}. Some others employ simple generative modules to map representations of available images to those of the missing ones \cite{MPLMM,MOME+}. The objective of implicit representation learning is to improve the performance of downstream tasks. It may sacrifice the fidelity of the synthesised images. Overall, both paradigms require further exploration of the mutual benefits between generation and understanding.

The recent emergence of unified multimodal models for generation and understanding offers a promising solution \cite{BAGEL,Mogao,Emu3}. By tokenising and modeling both visual and textual data within a single, autoregressive decoder-only architecture, the unified models natively intertwine the two tasks of generation and understanding. This enables multimodal interleaved reasoning, such as thinking with generated images \cite{UniCoT}, where the generation process enhances the chain-of-thought reasoning. However, developing unified models for medical domain faces three critical obstacles. First, there is a lack of interleaved and multimodal medical dataset for training a unified model. How to design tasks that effectively leverage the generation ability for multimodal understanding is still under-explored. Furthermore, there is a domain gap between natural and medical image-text pairs. Models trained on natural images are ill-equipped with specialist knowledge to understand medical images and learn clinically meaningful representations \cite{nath2025vila}. Finally, interleaved reasoning relies on intermediate generation results as a context to perform subsequent reasoning, which may lead to error accumulation, compromising the final model performance \cite{schmidt2019generalization}.

In this work, we propose UniBrain, a novel multimodal model to perform brain image analysis with missing MRI modalities. As a unified model, UniBrain simultaneously improves both generation fidelity and diagnostic accuracy through a multimodal autoregressive process. 
To overcome the aforementioned obstacles, we introduce three core contributions: 1) We design a novel framework for brain MRI diagnosis, which imputes missing MRI modalities as intermediate steps before formulating the final understanding output, enabling reasoning with imputed images. 2) % To overcome the scarcity of multimodal medical training data, w
We introduce a self-adaption strategy that leverages the visual embeddings of the model to reconstruct input medical images, which reduces the need for medical reports with fine-grained visual descriptions. 3) To mitigate accumulated errors in long-context autoregressive generation, we implement a dynamic hidden state mechanism that enforces the model to be aware of the self-generated artefacts during training, improving the robustness of autoregressive inference.

\begin{figure}[t]
\includegraphics[width=\textwidth]{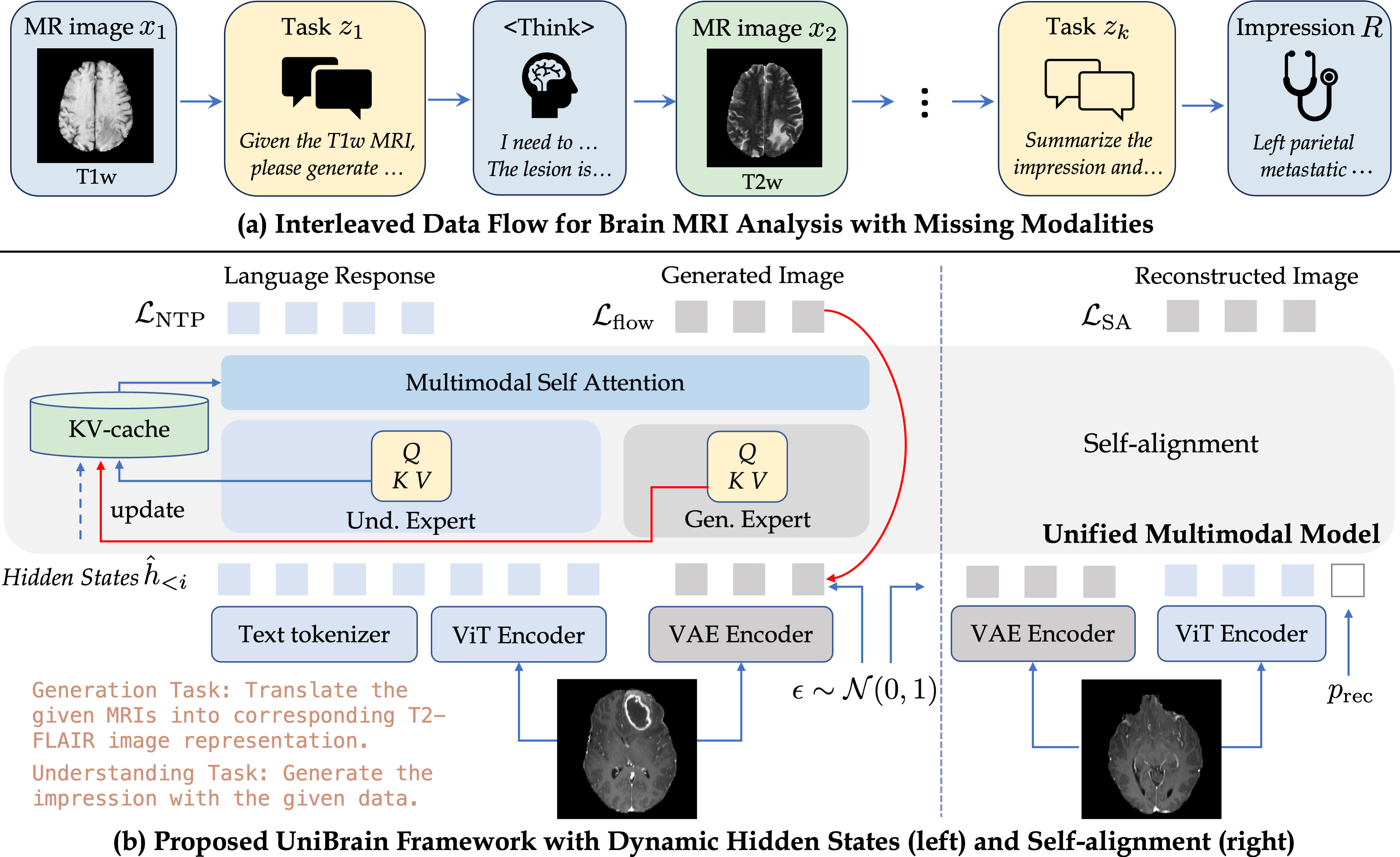}
\caption{Overall framework of UniBrain. (a) We construct a sequential, interleaved data flow for brain MRI analysis during training. 
(b) The framework incorporates a unified multimodal model for image generation and understanding. Dynamic hidden states via training-time KV-cache are designed to mitigate exposure bias for long-context multimodal reasoning. A self-alignment loss conditions the generation expert on dense embeddings extracted from the ViT for image reconstruction. 
} \label{overall}
\end{figure}

\section{Methodology}
\subsection{Preliminary of Unified Multimodal Model}
UniBrain is built upon BAGEL \cite{BAGEL}, a unified model for joint vision-language understanding and generation. Employing a Mixture-of-Transformer-Experts (MoT) architecture \cite{MOT}, BAGEL comprises two Transformer experts, respectively dedicated to understanding and generation. Given multimodal input, text is encoded by a text tokeniser, whereas images are encoded into visual tokens by two encoders, an understanding-oriented ViT and a generation-oriented VAE. The two experts operate over shared text and visual tokens via multimodal self-attention, which allows long-range interactions between the tokens. On the output side, BAGEL predicts the next token for language response and visual tokens for image generation via rectified flow \cite{BAGEL}. UniBrain adapts BAGEL to brain MRI analysis in three aspects, as explained in the following sections. %While being an effective architecture for unified modeling in general domain, BAGEL is not originally designed for modality imputation or medical image understanding, posing both the task and domain gaps for MRI analysis with missing modalities.

\subsection{Unified Brain MRI Modality Imputation and Understanding}
% Despite being pre-trained on large-scale interleaved text, image, and video data, the vanilla BAGEL architecture is primarily optimised for general-domain single-image manipulation or sequential video dynamics. However, in clinical neuroimaging, multi-parametric MRI (e.g., T1-weighted, T2-weighted, FLAIR) is the standard of care, as varying sequences offer complementary information on tissue morphology essential for comprehensive brain region analysis. To fully support diverse scanning protocols and bridge the gap between general-domain interleaved data and clinical reasoning, we construct a novel sequential, description-enriched medical modality flow.
We formulate brain MRI missing modality imputation and image understanding (\emph{e.g.} disease diagnosis) as an autoregressive sequence modeling problem. We denote the images of $k$ MRI modalities as $X{=}\{x_i\}_{i=1}^k$, the corresponding radiological findings for each modality as $F{=}\{f_i\}_{i=1}^k$, and the impression as $R$ which contains global description and diagnostic results. As illustrated in Fig.~\ref{overall}(a), given input image $x_1$, a multimodal model $\pi_{\theta}$, parameterised by $\theta$, iteratively executes reasoning-enriched generation tasks $Z_g{=}\{z_i\}_{i=1}^{k-1}$ and finally performs the understanding task $Z_u{=}z_k$. The overall interleaved data flow is constructed as $\{x_1, z_1, f_1, x_2, \dots, z_k, R\}$. The model contains hidden state $h_i$, which are tokenised images or text. 
It is trained to predict its next state based on previous accumulated multimodal context: $\hat{h}_i{=}\pi_{\theta} (x_1, z_i, h_{<i})$, where ${h}_{<i}{=}\{h_1, h_2, \dots, {h}_{i-1}\}$ denotes the history of hidden states.  
For generation task $z_i \in Z_g$, the model is instructed to perform the thinking procedure before generating the target image $x_{i+1}$. This thinking procedure consists of two components: 1) enriched description of the current objective $z_i$
(\emph{e.g.}, "\textit{I need to translate the given T1 and FLAIR images into the corresponding T2-weighted MRI}".), and
2) the radiological findings ($f_{\leq i}$) associated with the already known (ground-truth or synthesised) input modalities. For the final understanding task $z_k$, the model leverages the full accumulated sequence to produce the overall impression $R$. 

%In this paradigm, <Modality 0> represents the available ground-truth MRI sequences, which serve as the foundational visual context. Driven by a clinical <Instruction>, the model autoregressively synthesises the missing sequences (<Modality 1>, <Modality 2>), guided by a generalised causal attention mechanism that conditions generation on the preceding multimodal context. To explicitly bridge the visual-semantic gap during this cross-modality generation, we enrich the data flow with reasoning-augmented <Thinking> steps prior to each generation output. 
% Inspired by recent advances in chain-of-thought generation, our UFM is trained to formulate a conceptual and clinical plan before synthesising a missing sequence. Specifically, the <Thinking> process is structured to 

\subsection{Fine-grained MRI Representation with Self-alignment}
One challenge in adapting general-purpose unified models to medical imaging is to bridge the gap between medical imaging modalities and free-form texts.
% While the interleaved modality flow described above provides the structural scaffolding for clinical reasoning, a fundamental challenge remains: bridging the severe domain gap between natural images where BAGEL are predominantly pre-trained on and the complex, fine-grained anatomical textures of brain MRIs.
Conventional vision-language alignment relies heavily on abundant paired natural image-text data, while paired medical image-text data are extremely sparse. In addition, radiological text such as "\textit{hyperintense lesion in the right parieto-occipital lobe with ill-defined margins}" only provides a high-level description of a brain MRI scan and does not provide pixel-level supervisory signals. This leads to the discrepancy between the ViT tokens used for high-level understanding and VAE tokens used for pixel-level MRI generation. %, such as boundaries, textures, or precise tissue contrasts etc, which are essential for accurate medical image generation. 

To overcome the limitation, we propose a self-alignment (SA) strategy that leverages intrinsic supervisory signals from the images themselves. Drawing inspiration from \cite{RecA}, we postulate that visual embeddings extracted from a model's visual understanding encoder may contain richer semantic representations of an image than what radiological text caption could capture. Using these visual embeddings as prompts, a unified model can be trained to reconstruct the original image.
%Rather than applying RecA merely as a general-domain post-training enhancement, we repurpose this mechanism specifically as a domain adaptation strategy for multi-sequence MRI.
In detail, we condition the generation expert on the dense visual embeddings of image $x_i$ extracted from the ViT encoder.
As the proposed model performs visual token prediction in the latent space, the self-supervised image reconstruction objective is formulated accordingly. The VAE encoder projects $x_i$ into a latent representation $e_i = E_{\text{VAE}}(x_i)$, while the ViT encoder extracts the dense visual embeddings $w_i = E_{\text{ViT}}(x_i)$. The training objective for self-alignment is to predict the velocity in the latent space of the target MRI modality, conditioned on the visual embeddings $w_i$ and the text prompt $p_\text{rec}$ to instruct reconstruction. The self-alignment loss is formulated as:
\begin{equation}
\mathcal{L}_\text{SA} = \mathbb{E}_{t\sim\mathcal{U}[0,1],\epsilon\sim\mathcal{N}(0,I)}[|| V_\theta(e_i^t, t, w_i, p_\text{rec})-(\epsilon - e_i) ||^2], 
\end{equation}
where $e_i^t=(1-t)e_i+t\epsilon$ represents the interpolated latent  between the clean latent $e_i$ and the noise sample $\epsilon$. $V$ is the velocity prediction network \cite{esser2024scaling} parameterised by the generation expert.
As the training of $\mathcal{L}_\text{SA}$ requires only the image itself, the model learns from both highly pathological images and healthy images, alleviating the need for detailed visual descriptions of these images. 

\subsection{Bridging the Training-test Gap via Dynamic Hidden States}
\label{sec:hidden states}
Another challenge for unified models is the training-test domain gap, commonly referred to as the exposure bias \cite{ExposureBias}. This gap originated from the fact that autoregressive inference is based on the previous predicted states $\hat{h}_{<i}$, which are generated and thus with a bias from the ground-truth states $h_{<i}$ (used during training). Although BAGEL implements a diffusion forcing strategy to alleviate the exposure bias, it operates in the ground-truth noisy latent space, leading to accumulated errors for the context $(x_1, z_i, \hat{h}_{<i})$ when $i$ increases. 

To bridge this gap, we propose a dynamic hidden state (DHS) mechanism that enforces the model to condition on its own visual predictions during training. Due to the long-context scenario that demands massive computation, we leverage a training-time KV cache mechanism~\cite{SelfForcing} to enable an affordable end-to-end autoregressive rollout of the constructed data flow.
We first process the deterministic, known information. The system instructions and input modality $x_1$ are encoded and tokenised, with their activations being stored in the KV cache. When the data flow reaches a generation token (\emph{i.e.}, tokenised $e_i^t$), we initiate a few-step (10 steps in our case) diffusion \cite{SelfForcing} in the VAE latent space to generate the current missing modality, where generation loss is computed on the exit timestep. This generation is conditioned on the previously KV cache containing history hidden states. Then, we project the self-generated visual tokens $\hat{e}_i$ back to the model's hidden state $\hat{h}_i$ and append it to the KV cache. Finally, the understanding task $z_k$ is processed using the fully accumulated KV cache. Because this cache now contains the self-generated dynamic intermediate states rather than the ground-truth static ones, the model is forced to provide accurate results despite the presence of its own generative artefacts.

Following the procedure, UniBrain is trained with a unified loss function modified with dynamic states. For text-related tasks (thinking and reporting), we employ the next-token prediction (NTP) loss for each hidden state:
\begin{equation}
    \mathcal{L}_\text{NTP} = -\sum_{n}\log p(h_{i,n}|\textbf{KV}_{< i}, h_{i,<n}; \theta),
\end{equation} where $h_{i,n}$ denotes the $n$-th token in its current state, $\textbf{KV}_{< i}$ denotes the KV cache for previous states. For visual generation, we apply flow matching on the latents:
\begin{equation}
    \mathcal{L}_\text{flow} = \mathbb{E}_{t \sim U[0,1],\epsilon \sim \mathcal{N}(0,I)}[||V_{\theta}(e_i^t, t, \textbf{KV}_{< i}) - (\epsilon - e_i)||^2],
\end{equation} where the model predicts the velocity for interpolated latent $e_i^t$ conditioned on the previous KV cache. We keep the bidirectional attention on visual tokens in current states. The overall loss can be formulated as:
\begin{equation}
    \mathcal{L} = \frac{1}{k}\sum_i (\mathcal{L}_{\text{SA}} + \mathcal{L}_{\text{NTP}} + \mathcal{L}_{\text{flow}}).
\end{equation}
During inference, each patient case provides $a$ available  modalities $\{x_i\}_{i=1}^a$ as inputs, while the remaining modalities $\{x_i\}_{i=a+1}^k$ are missing and need to be synthesised. We sequentially perform the generation tasks $\{z_i\}_{i=a}^{k-1},$ before performing the final understanding task $z_k$. For each task, the model predicts its next state based on the accumulated multimodal context: $\hat{h}_i{=}\pi_{\theta} (x_1, z_i, \hat{h}_{<i})$, where $\hat{h}_{<i}{=}\{h_1, \dots, h_{a}, \hat{h}_{a+1}, \dots, \hat{h}_{i-1}\}$ denotes the history of hidden states. %\cref{overall}~(a) illustrates the sequential tasks, where the overall interleaved data flow is constructed as $\{x_1, z_1, f_1, x_2, \dots, z_k, R\}$. For generation task $z_i$ ($i<k$), the model is instructed to perform the thinking procedure before generating the imputed target image $x_{i+1}$. This explicit chain-of-thought consists of two components: 1) enriched description of the current objective $z_i$ (\emph{e.g.}, "\textit{I need to translate the given T1 and FLAIR images into the corresponding T2-weighted MRI}".), and 2) the radiological findings ($f_{\leq i}$) associated with the already known (ground-truth or synthesised) input modalities. For the final understanding task $z_k$, the model leverages the fully accumulated sequence to produce the overall impression $R$. 

\section{Experiments}
\subsection{Data and Experiment Settings}
\subsubsection{Dataset} A public dataset, RadGenome-Brain MRI \cite{RadGenome}, is used for model training and evaluation, which contains 3,408 brain MRI scans acquired from multiple sites with paired radiology reports, covering six MRI modalities (T1, T2, FLAIR, DWI, ADC, and T1ce) and five neurological disease types (glioma, meningioma, metastasis, stroke, white matter hyperintensities (WMH)). The reports were written by senior radiologists, providing findings and impressions grounded in the annotated image regions \cite{RadGenome}. We follow the official setting to split the dataset into training ($n$=2,382), validation ($n$=338), and test sets ($n$=688).% \pdfcomment{There should be data split info, otherwise reviewer would be very critical...}

\subsubsection{Implementation Details}
UniBrain is built upon the BAGEL backbone with 14B parameters pretrained on large‑scale multimodal data \cite{BAGEL}.
%Following common practices, we utilise FlashAttention, fully sharded data parallel (FSDP), and mixed-precision training for better computational efficiency. 
% The unified model are optimised using the Adam optimiser with a learning rate of 2e-5. 
% We employed a linear warm-up learning rate schedule for training, increasing the learning rate from 0 to 2e-5 in the first 200 steps. 
To stabilise training, the model was trained with self-alignment for 2,000 steps, followed by 2,000 steps for non-KV cache training and 200 steps using the DHS mechanism. At each step, we sample a batch of brain MRI slices corresponding to annotated lesions and randomise the number and order of MRI modalities. The corresponding impressions and clinical findings, together with the images are packed into a sequence of 13,408 tokens for model training. Since 2D slices are used, we reformulate the 3D volumetric and size-related descriptions in the original reports and project them to 2D slice-specific descriptors. We randomly drop the text and visual tokens with the probability of 0.1 and 0.3, respectively. All experiments were performed on 8 Nvidia A100 GPUs.

\subsection{Results}
\subsubsection{Diagnosis and Report Generation Performance with Missing Modalities}
Table \ref{tab:progressive_missing}  reports the top-1 diagnostic accuracy (Top-1) \cite{bercea2025nova} and report generation quality (ROUGE-1 \cite{lin2004rouge}, RaTEScore \cite{zhao2024ratescore}) under different missing-modality scenarios, compared between different methods that handle missing modalities. While implicit representation learning methods (ShaSpec\cite{ShaSpec} and SimMLM \cite{SimMLM}) achieve reasonable diagnostic accuracy, they lack clinical interpretability. Explicit modality imputation methods (ResViT \cite{ResViT} and M2DN \cite{M2DN}) perform poorly in diagnostic accuracy. This is because the imputation methods are trained separately from the MLLMs, leading to misaligned features between imputed images and MLLMs. Instead, UniBrain overcomes this issue via unified modeling and outperforms other MLLMs. It achieves 74.47\% diagnostic accuracy in the highly challenging setting with only T1w modality available. With complete data, the diagnostic accuracy further increases to 82.06\%. %UniBrain consistently outperforms other MLLMs, including BAGEL \cite{BAGEL}, UniMedVL \cite{UniMedVL} and Lingshu \cite{Lingshu}, in terms of both diagnosis and report generation. 
% While our understanding-only variant (UniBrain Und.) establishes a strong multimodal baseline, integrating the unified diagnosis flow yields a further 5\% improvement, achieving 74.47\% Acc-1 in the highly challenging T1w-only setting.

\begin{figure}[tbp]
\includegraphics[width=\textwidth]{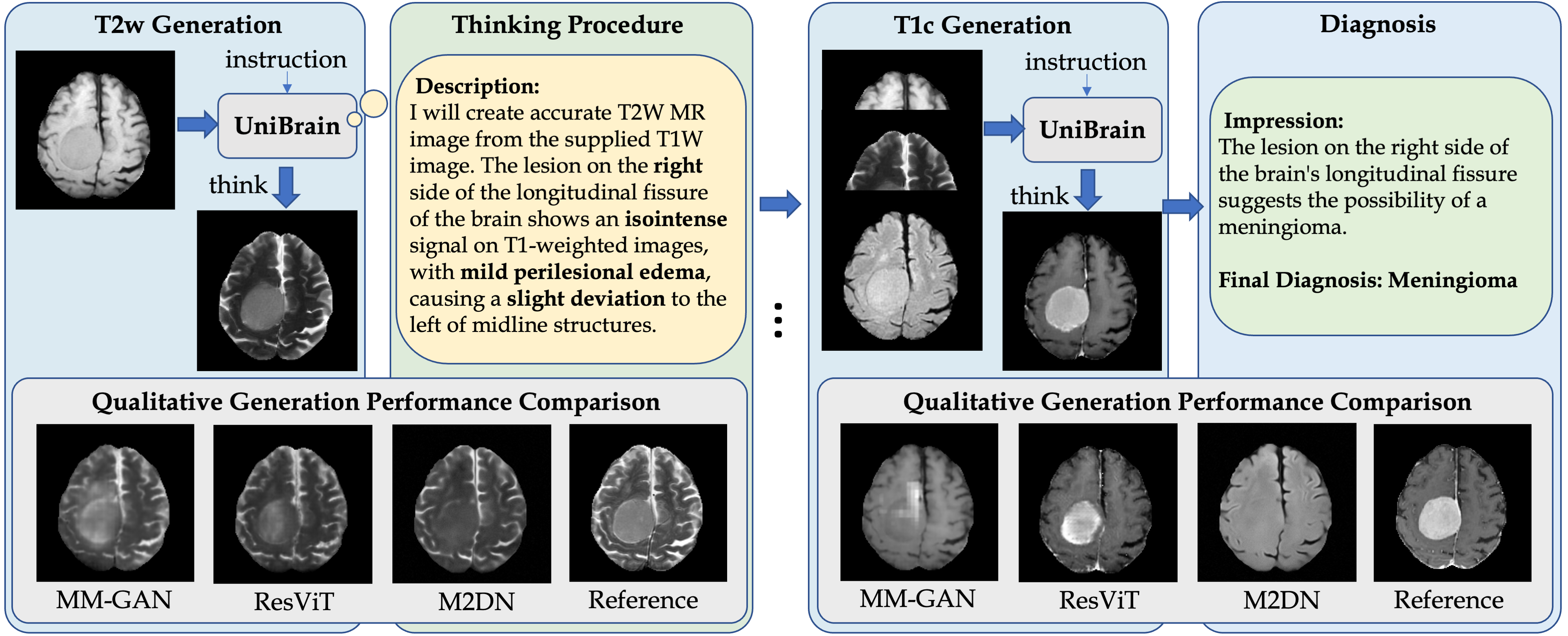}
\caption{Qualitative results for unified brain MRI analysis using UniBrain.
}
\label{results}
\end{figure}

\begin{table*}[t]
\centering
\caption{Quantitative results for diagnosis and report generation with missing MRIs. Und. means the understanding-only counterparts of the model.}
\label{tab:progressive_missing}
\resizebox{\textwidth}{!}{
\begin{tabular}{lcccccccccccc}
\toprule
\multirow{2}{*}{\textbf{Methods}} & \multicolumn{3}{c}{\textbf{T1w only}} & \multicolumn{3}{c}{\textbf{T1w + T2w}} & \multicolumn{3}{c}{\textbf{T1w + T2w + T2f}} & \multicolumn{3}{c}{\textbf{Complete Data}} \\
\cmidrule(lr){2-4} \cmidrule(lr){5-7} \cmidrule(lr){8-10} \cmidrule(lr){11-13}
& \textbf{Top-1} & \textbf{ROUGE} & \textbf{RaTE} & \textbf{Top-1} & \textbf{ROUGE} & \textbf{RaTE} & \textbf{Top-1} & \textbf{ROUGE} & \textbf{RaTE} & \textbf{Top-1} & \textbf{ROUGE} & \textbf{RaTE} \\
\midrule
\multicolumn{13}{l}{\textit{Implicit Representation Learning}} \\
ShaSpec \cite{ShaSpec} & 63.61 & - & - & 71.98 & - & - & 75.26 & - & - & 77.19 & - & - \\
SimMLM \cite{SimMLM} & 65.98 & - & - & 74.47 & - & - & 76.60 & - & - & 78.72 & - & - \\
\midrule
\multicolumn{13}{l}{\textit{Explicit Modality Imputation}} \\
ResViT \cite{ResViT} + Lingshu \cite{Lingshu} & 24.82 & 20.79 & 37.83 & 31.21 & 19.06 & 34.83 & 37.59 & 19.41 & 37.08 & - & - & -\\
ResViT \cite{ResViT} + UniBrain & 59.57 & 35.10 & 56.62 & 67.38 & 36.49 & 59.18 & 73.05 & 38.28 & 59.88 & - & - & -\\
M2DN \cite{M2DN} + UniBrain & 56.03 & 33.93 & 57.76 & 75.18 & 36.38 & 60.08 & 76.60 & 38.36 & 60.98 & - &- & - \\
\midrule
\multicolumn{13}{l}{\textit{Multimodal Large Language Models}} \\
BAGEL \cite{BAGEL} & 11.35 & 17.02 & 44.37 & 17.02 & 15.43 & 41.08 & 22.70 & 16.12 & 42.35 & 28.37 & 17.76 & 43.19 \\
UniMedVL \cite{UniMedVL} & 29.79 & 13.90 & 42.58 & 30.50 & 14.73 & 41.84 & 32.62 & 13.81 & 43.58 & 38.30 & 13.70 & 43.97 \\
Lingshu \cite{Lingshu} & 21.99 & 18.08 & 36.91 & 24.82 & 19.49 & 35.55 & 29.79 & 18.13 & 34.32 & 41.13 & 20.26 & 37.29 \\
\textbf{UniBrain Und.} & 69.50 & \textbf{37.35} & 61.50 & 73.05 & 35.94 & 60.48 & 76.60 & 38.05 & 61.04 & \textbf{82.06} & \textbf{38.94} & \textbf{61.62} \\
\textbf{UniBrain} & \textbf{74.47} & 36.93 & \textbf{61.57} & \textbf{76.60} & \textbf{38.23} & \textbf{61.24} & \textbf{78.01} & \textbf{38.68} & \textbf{61.90} & \textbf{82.06} & \textbf{38.94} & \textbf{61.62} \\
\bottomrule
\end{tabular}
}
\end{table*}

\subsubsection{Missing Modality Imputation Fidelity and Diagnostic Usability}
Table \ref{tab:imputation_fidelity}  reports the fidelity of the generated missing modalities in terms of peak signal-to-noise ratio (PSNR), structural similarity (SSIM) and the usability for downstream diagnostic task in terms of Acc-1, evaluated via the understanding expert of UniBrain. Compared to a medical multi-modal model UniMedVL \cite{UniMedVL}, UniBrain achieves better performance in both imputation fidelity and diagnostic usability. Compared to strong generative model baselines, which are image-only models specifically designed for imputation, UniBrain achieves similar or slightly lower performance in imputation fidelity. However, it achieves much higher diagnostic usability in terms of Acc-1. Also, we notice that if UniBrain is ensembled with five random seeds, it can outperform specifically-designed imputation models in terms of low-level fidelity metrics (PSNR and SSIM) while maintaining superior diagnostic performance.
% We can find in Table \ref{tab:imputation_fidelity} that image translation metrics may not faithfully reflect the diagnostic usability of generated image: UniBrain significantly outperforms the counterparts in downstream diagnostic accuracy, while the generation metrics are not among the best. We attribute this to the fact that other methods produce blurry results that cannot faithfully capture the anomaly regions, as showcased in Fig. \ref{results}. By sampling for 5 random seeds and averaging the results, most metrics can be greatly improved, despite the longer inference time.

\begin{table*}[t]
\centering
\caption{Quantitative evaluation of missing modality imputation. * means we ensemble results sampled with 5 random seeds.}
\label{tab:imputation_fidelity}
\resizebox{\textwidth}{!}{
\begin{tabular}{llccccccccc}
\toprule
\multirow{2}{*}{\textbf{Methods}} & \multirow{2}{*}{\textbf{Architecture}} & \multicolumn{3}{c}{\textbf{T1w $\rightarrow$ T2w}} & \multicolumn{3}{c}{\textbf{T1w, T2w $\rightarrow$ T2f}} & \multicolumn{3}{c}{\textbf{T1w, T2w, T2f $\rightarrow$ T1c}} \\
\cmidrule(lr){3-5} \cmidrule(lr){6-8} \cmidrule(lr){9-11}
& & \textbf{PSNR} & \textbf{SSIM} & \textbf{Top-1} & \textbf{PSNR} & \textbf{SSIM} & \textbf{Top-1} & \textbf{PSNR} & \textbf{SSIM} & \textbf{Top-1} \\
\midrule
\multicolumn{11}{l}{\textit{Generative Baselines}} \\
MM-GAN \cite{MMGAN} & GAN & 23.08 & 0.8774 & 56.74 & 23.32 & 0.8719 & 56.03 & 23.40 & 0.8702 & 60.19 \\
ResViT \cite{ResViT} & GAN & 22.81 & 0.8751 & 57.45 & 23.13 & 0.8740 & 67.38 & 23.00 & \textbf{0.8735} & 61.70 \\
M2DN \cite{M2DN} & Diffusion & 22.79 & 0.8019 & 51.06 & 22.46 & 0.7764 & 51.06 & 22.05 & 0.7820 & 61.70 \\
\midrule
\multicolumn{11}{l}{\textit{Unified Multimodal Models}} \\
UniMedVL \cite{UniMedVL} & Unified Model & 19.82 & 0.7230 & 56.03 & 19.96 & 0.6608 & 63.12 & 21.53 & 0.6599 & 56.74 \\
\textbf{UniBrain} & \textbf{Unified Model} & 22.23 & 0.8637 & \textbf{68.09} & 22.58 & 0.8613 & 67.38 & 22.26 & 0.8617 & 74.47 \\
\textbf{UniBrain*} & \textbf{Unified Model} & \textbf{23.43} & \textbf{0.8850} & 63.83 & \textbf{23.49} & \textbf{0.8760} & \textbf{68.08} & \textbf{23.52} & 0.8725 & \textbf{76.60} \\
\bottomrule
\end{tabular}
}
\end{table*}

\begin{table*}[t]
\centering
\caption{Ablation studies for UniBrain.}
\label{tab:ablation}
\resizebox{\textwidth}{!}{
\begin{tabular}{lcccccccccc}
\toprule
\multirow{2}{*}{\textbf{Model}} & \multicolumn{3}{c}{\textbf{Components}} & \multicolumn{3}{c}{\textbf{Understanding (T1w only)}} & \multicolumn{3}{c}{\textbf{Generation (T1w$\rightarrow$ ... $\rightarrow$T1c)}} \\
\cmidrule(lr){2-4} \cmidrule(lr){5-7} \cmidrule(lr){8-10}
& \textbf{Unified} & \textbf{SA} & \textbf{DHS} & \textbf{Acc-1} & \textbf{ROUGE} & \textbf{RaTEScore} & \textbf{PSNR} & \textbf{SSIM} & \textbf{Top-1} \\
\midrule
A & & & & 70.05 & 35.71 & 60.12 & - & - & - \\
B & \checkmark & & & \textbf{75.11} & 36.35 & 60.52 & 21.28 & 0.8329 & 70.12 \\
C & \checkmark & \checkmark & & 73.76 & 35.34 & 59.23 & 22.09 & 0.8456 & 74.03 \\
\textbf{UniBrain} & \checkmark & \checkmark & \checkmark & 74.47 & \textbf{36.93} & \textbf{61.57} & \textbf{22.47} & \textbf{0.8519} & \textbf{76.60} \\
\bottomrule
\end{tabular}
}
\end{table*}

\subsubsection{Ablation Studies}
Table \ref{tab:ablation}  reports quantitative results for ablation studies to evaluate the contributions of each component in UniBrain. We start with a vanilla BAGEL architecture (A) and perform the understanding task, serving as a strong baseline. Then, we introduce interleaved data for unifed modeling (B) and enable understanding with imputed modalities, which greatly improve the diagnosis performance (by 5.06\%). Furthermore, we impose self-alignment (SA) strategy (C) to enrich fine-grained representation, which benefits the generation quality (+0.81 in PSNR) while slightly disturbing the diagnosis accuracy. The final UniBrain model introduces the dynamic hidden state (DHS) mechanism, achieving optimal overall performance for both generation and understanding.

\section{Conclusion}
In this paper, we introduce UniBrain, a novel unified multimodal model designed to perform multimodal brain MRI analysis with potential missing modalities. UniBrain adapts an autoregressive architecture to represent fine-grained MRI details via self-alignment and conduct multimodal reasoning with the dynamic hidden state mechanism.
Extensive evaluations demonstrate that UniBrain achieves strong performance in terms of modality imputation fidelity, report generation quality, and disease diagnosis accuracy. % under varying degrees of data incompleteness. 
Future work will explore improving computational efficiency (currently requiring over 32GB GPU memory for inference), involving radiologists in model evaluation for diverse disease cohorts, adapting the model from 2D to 3D, and extending data modalities to further enrich the diagnostic context and advance towards multimodal AI for healthcare.

\begin{credits}
\subsubsection{\ackname} This work was supported by Imperial College London President's PhD Scholarship, EPSRC CVD-Net Programme Grant (EP/Z531297/1) and BHF New Horizons Grant (NH/F/23/70013). A.K. was supported by the EPSRC Doctoral Prize Fellowship. T.X. is supported through the Imperial College London UKRI Impact Acceleration Account EP/X52556X/1.

\subsubsection{\discintname}
The authors have no competing interests to declare.
\end{credits}

\bibliographystyle{splncs04}
\bibliography{Paper-2626}

@inproceedings{bercea2025nova,
  title={{NOVA: A benchmark for rare anomaly localization and clinical reasoning in brain MRI}},
  author={Bercea, Cosmin I and Li, Jun and Raffler, Philipp and Riedel, Evamaria Olga and Schmitzer, Lena and others},
  booktitle={Neural Information Processing Systems},
  year={2025}
}

@inproceedings{shen2019brain,
  title={{Brain tumor segmentation on MRI with missing modalities}},
  author={Shen, Yan and Gao, Mingchen},
  booktitle={International Conference on Information Processing in Medical Imaging},
  year={2019}
}

@inproceedings{chen2024towards,
  title={{Towards injecting medical visual knowledge into multimodal LLMs at scale}},
  author={Chen, Junying and Gui, Chi and Ouyang, Ruyi and Gao, Anningzhe and Chen, Shunian and others},
  booktitle={Conference on Empirical Methods in Natural Language Processing},
  year={2024}
}

@article{yin2024survey,
  title={A survey on multimodal large language models},
  author={Yin, Shukang and Fu, Chaoyou and Zhao, Sirui and Li, Ke and Sun, Xing and Xu, Tong and Chen, Enhong},
  journal={National Science Review},
  year={2024},
  publisher={Oxford University Press}
}

@article{alsaad2024multimodal,
  title={Multimodal large language models in health care: applications, challenges, and future outlook},
  author={AlSaad, Rawan and Abd-Alrazaq, Alaa and Boughorbel, Sabri and Ahmed, Arfan and Renault, Max-Antoine and Damseh, Rafat and Sheikh, Javaid},
  journal={Journal of Medical Internet Research},
  volume={26},
  pages={e59505},
  year={2024},
  publisher={JMIR Publications Toronto, Canada}
}

@article{UniMedVL,
  title={{UniMedVL: Unifying medical multimodal understanding and generation through observation-knowledge-analysis}},
  author={Ning, Junzhi and Li, Wei and Tang, Cheng and Lin, Jiashi and Ma, Chenglong and others},
  journal={arXiv preprint arXiv:2510.15710},
  year={2025}
}

@inproceedings{
UniCoT,
title={{Uni-CoT: Towards unified chain-of-thought reasoning across text and vision}},
author={Qin, Luozheng and Gong, Jia and Sun, Yuqing and Li, Tianjiao and Yang, Mengping and others},
booktitle={International Conference on Learning Representations},
year={2026}
}

@inproceedings{nath2025vila,
  title={{Vila-m3: Enhancing vision-language models with medical expert knowledge}},
  author={Nath, Vishwesh and Li, Wenqi and Yang, Dong and Myronenko, Andriy and Zheng, Mingxin and others},
  booktitle={Proceedings of the Computer Vision and Pattern Recognition},
  year={2025}
}

@inproceedings{schmidt2019generalization,
  title={Generalization in generation: A closer look at exposure bias},
  author={Schmidt, Florian},
  booktitle={Proceedings of the 3rd Workshop on Neural Generation and Translation},
  year={2019}
}

@article{
wu2026deep,
title={Deep Multimodal Learning with Missing Modality: A Survey},
author={Renjie Wu and Hu Wang and Hsiang-Ting Chen and Gustavo Carneiro},
journal={Transactions on Machine Learning Research},
issn={2835-8856},
year={2026}
}

@article{UniCOAL,
  title={{Uni-COAL: A unified framework for cross-modality synthesis and super-resolution of MR images}},
  author={Song, Zhiyun and Qi, Zengxin and Wang, Xin and Zhao, Xiangyu and Shen, Zhenrong and others},
  journal={Expert Systems with Applications},
  volume={270},
  pages={126241},
  year={2025},
  publisher={Elsevier}
}

@article{ResViT,
  title={{ResViT: Residual vision transformers for multimodal medical image synthesis}},
  author={Dalmaz, Onat and Yurt, Mahmut and {\c{C}}ukur, Tolga},
  journal={IEEE Transactions on Medical Imaging},
  volume={41},
  number={10},
  year={2022},
  publisher={IEEE}
}

@article{rassmann2026regression,
  title={Regression is all you need for medical image translation},
  author={Rassmann, Sebastian and K{\"u}gler, David and Ewert, Christian and Reuter, Martin},
  journal={IEEE Transactions on Medical Imaging},
  year={2026},
  publisher={IEEE}
}

@article{zhao2024deep,
  title={Deep multimodal data fusion},
  author={Zhao, Fei and Zhang, Chengcui and Geng, Baocheng},
  journal={ACM Computing Surveys},
  volume={56},
  number={9},
  pages={1--36},
  year={2024},
  publisher={ACM New York, NY}
}

@article{rahimpour2021cross,
  title={{Cross-modal distillation to improve MRI-based brain tumor segmentation with missing MRI sequences}},
  author={Rahimpour, Masoomeh and Bertels, Jeroen and Radwan, Ahmed and Vandermeulen, Henri and Sunaert, Stefan and others},
  journal={IEEE Transactions on Biomedical Engineering},
  volume={69},
  number={7},
  year={2021},
  publisher={IEEE}
}

@inproceedings{MPLMM,
  title={Multimodal prompt learning with missing modalities for sentiment analysis and emotion recognition},
  author={Guo, Zirun and Jin, Tao and Zhao, Zhou},
  booktitle={Proceedings of the Association for Computational Linguistics},
  year={2024}
}

@ARTICLE{MOME+,
  author={Zhang, Xinru and Ou, Ni and Doga Basaran, Berke and Visentin, Marco and Qiao, Mengyun and others},
  journal={IEEE Transactions on Medical Imaging}, 
  title={{A foundation model for lesion segmentation on brain MRI with mixture of modality experts}}, 
  year={2025},
  volume={44},
  number={6}
  }

@article{Emu3,
  title={Multimodal learning with next-token prediction for large multimodal models},
  author={Wang, Xinlong and Cui, Yufeng and Wang, Jinsheng and Zhang, Fan and Wang, Yueze and Zhang, Xiaosong and Luo, Zhengxiong and Sun, Quan and Li, Zhen and Wang, Yuqi and others},
  journal={Nature},
  pages={1--7},
  year={2026},
  publisher={Nature Publishing Group UK London}
}

@article{Mogao,
  title={{Mogao: An omni foundation model for interleaved multi-modal generation}},
  author={Liao, Chao and Liu, Liyang and Wang, Xun and Luo, Zhengxiong and Zhang, Xinyu and others},
  journal={arXiv preprint arXiv:2505.05472},
  year={2025}
}

@article{BAGEL,
  title   = {Emerging Properties in Unified Multimodal Pretraining},
  author  = {Deng, Chaorui and Zhu, Deyao and Li, Kunchang and Gou, Chenhui and Li, Feng and others},
  journal = {arXiv preprint arXiv:2505.14683},
  year    = {2025}
}

@article{
  MOT,
  title={{Mixture-of-Transformers: A sparse and scalable architecture for multi-modal foundation models}},
  author={Weixin Liang and Lili Yu and Liang Luo and Srini Iyer and Ning Dong and others},
  journal={Transactions on Machine Learning Research},
  issn={2835-8856},
  year={2025}
}

@inproceedings{
RecA,
title={Reconstruction Alignment Improves Unified Multimodal Models},
author={Ji Xie and Trevor Darrell and Luke Zettlemoyer and XuDong Wang},
booktitle={International Conference on Learning Representations},
year={2026},
}

@inproceedings{
SelfForcing,
title={Self Forcing: Bridging the Train-Test Gap in Autoregressive Video Diffusion},
author={Xun Huang and Zhengqi Li and Guande He and Mingyuan Zhou and Eli Shechtman},
booktitle={Neural Information Processing Systems},
year={2025}
}

@ARTICLE{RadGenome,
  author={Lei, Jiayu and Zhang, Xiaoman and Wu, Chaoyi and Dai, Lisong and Zhang, Ya and others},
  journal={IEEE Journal of Biomedical and Health Informatics}, 
  title={{Interpretable brain MRI report generation anchored by lesion topography}}, 
  year={2025}
  }

@inproceedings{ShaSpec,
  title={Multi-modal learning with missing modality via shared-specific feature modelling},
  author={Wang, Hu and Chen, Yuanhong and Ma, Congbo and Avery, Jodie and Hull, Louise and Carneiro, Gustavo},
  booktitle={IEEE/CVF Conference on Computer Vision and Pattern Recognition},
  year={2023}
}

@inproceedings{SimMLM,
  title={{SimMLM: A simple framework for multi-modal learning with missing modality}},
  author={Li, Sijie and Chen, Chen and Han, Jungong},
  booktitle={International Conference on Computer Vision},
  year={2025}
}

@article{Lingshu,
  title={Lingshu: A generalist foundation model for unified multimodal medical understanding and reasoning},
  author={Xu, Weiwen and Chan, Hou Pong and Li, Long and Aljunied, Mahani and Yuan, Ruifeng and  others},
  journal={arXiv preprint arXiv:2506.07044},
  year={2025}
}

@ARTICLE{MMGAN,
  author={A. {Sharma} and G. {Hamarneh}},
  journal={IEEE Transactions on Medical Imaging}, 
  title={{Missing MRI pulse sequence synthesis using multi-modal generative adversarial network}}, 
  year={2020}
  }

@ARTICLE{M2DN,
  author={Meng, Xiangxi and Sun, Kaicong and Xu, Jun and He, Xuming and Shen, Dinggang},
  journal={IEEE Transactions on Medical Imaging}, 
  title={Multi-modal modality-masked diffusion network for brain MRI synthesis with random modality missing}, 
  year={2024},
  volume={43},
  number={7},
  pages={2587-2598},}

@inproceedings{lin2004rouge,
  title={Rouge: A package for automatic evaluation of summaries},
  author={Lin, Chin-Yew},
  booktitle={ACL Workshop on Text Summarization Branches Out},
  year={2004}
}

@inproceedings{zhao2024ratescore,
  title={{RaTEScore: A metric for radiology report generation}},
  author={Zhao, Weike and Wu, Chaoyi and Zhang, Xiaoman and Zhang, Ya and Wang, Yanfeng and Xie, Weidi},
  booktitle={Conference on Empirical Methods in Natural Language Processing},
  year={2024}
}

@inproceedings{
esser2024scaling,
title={Scaling rectified flow transformers for high-resolution image synthesis},
author={Patrick Esser and Sumith Kulal and Andreas Blattmann and Rahim Entezari and Jonas M{\"u}ller and Harry Saini and Yam Levi and Dominik Lorenz and Axel Sauer and Frederic Boesel and Dustin Podell and Tim Dockhorn and Zion English and Robin Rombach},
booktitle={International Conference on Machine Learning},
year={2024},
}

@inproceedings{
ExposureBias,
title={Sequence level training with recurrent neural networks},
author={Ranzato, Marc'Aurelio and Chopra, Sumit and Auli, Michael and Zaremba, Wojciech},
booktitle={International Conference on Learning Representations},
year={2016},
}
\end{document}